\documentclass[letterpaper]{article} 
\usepackage[preprint]{aaai2027}  
\usepackage[hyphens]{url}  
\usepackage{graphicx} 
\usepackage{natbib}  
\usepackage{caption} 
\usepackage{algorithm}
\usepackage{algorithmic}

\usepackage{newfloat}
\usepackage{listings}
\DeclareCaptionStyle{ruled}{labelfont=normalfont,labelsep=colon,strut=off} 
\floatstyle{ruled}
\newfloat{listing}{tb}{lst}{}
\floatname{listing}{Listing}

\usepackage{booktabs}

\usepackage{amsmath}
\usepackage{amssymb}
\usepackage{multirow}

\title{HyBRA-EEG: Hyperbolic Global-Local Brain-Region Aggregation for Self-Supervised EEG Representation Learning}
\author{
    Xin Liu\equalcontrib,
    Feng Lu\equalcontrib,
    Wei Li\equalcontrib
}
\affiliations{}

\begin{document}

\maketitle

\begin{abstract}
Electroencephalography (EEG) representation learning is challenged by noisy signals, inter-subject variability, and heterogeneous electrode montages in different datasets and recording settings. Recent self-supervised EEG pretraining methods improve transferability by converting heterogeneous recordings into unified token or sequence representations, but channel-aware designs may still encode montage- or dataset-specific regularities. We propose HyBRA, a self-supervised framework based on hyperbolic global-local brain-region aggregation. HyBRA maps heterogeneous channel observations into a fixed regional token space by combining electrode coordinates, region priors, and feature-dependent hyperbolic distances. A channel-evidence injection branch preserves fine-grained discriminative cues during aggregation. The resulting tokens are processed by a Region-Time Transformer and pretrained with masked reconstruction of latent features and region-level signals. HyBRA outperforms the compared methods in average balanced accuracy on six benchmark datasets. Under a controlled six-fold evaluation on OpenBMI-MI, HyBRA matches CSBrain in balanced accuracy, improves AUROC by 1.84 percentage points and AUPRC by 5.23 percentage points, and uses 96.8\% fewer parameters. Matched comparisons on TUAB and TUEV show higher balanced accuracy for hyperbolic matching than for Euclidean distance or cosine similarity.
\end{abstract}

\section{Introduction}
\label{sec:introduction}

Electroencephalography (EEG) provides a non-invasive view of brain activity and supports applications in clinical EEG analysis and
brain--computer interfaces~\cite{daly2008brain,cohen2017where}. Transfer between datasets remains difficult under subject and acquisition variability, local waveform distortion, and montage-dependent changes in the spatial sampling of scalp potentials~\cite{jurcak2007systems,wu2022transfer}. Recent self-supervised models, including BIOT~\cite{yang2023biot}, EEGPT~\cite{wang2024eegpt}, LaBraM~\cite{jiang2024labram}, CBraMod~\cite{wang2025cbramod}, and CSBrain~\cite{zhou2025csbrain}, learn transferable representations from large EEG corpora while accommodating heterogeneous recording formats. These designs improve compatibility with different channel numbers and layouts. Their learned spatial organization may remain tied to dataset-specific channel identities and montages.

\begin{figure}[t]
    \centering
    \includegraphics[width=\columnwidth]{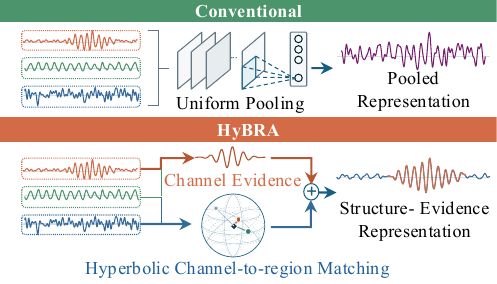}
    \caption{Conceptual comparison between a schematic uniform regional aggregation baseline and HyBRA. The baseline assigns equal importance to channels within each region and may attenuate informative activity. HyBRA combines feature-dependent hyperbolic channel-to-region matching with a direct channel-evidence pathway to form fused region tokens. Waveforms are schematic.}
    \label{fig:idea}
\end{figure}

Montage sensitivity persists when spatial meaning remains attached to channel identity. A channel token from one dataset may have no direct counterpart in another, and similarly named electrodes may exhibit different signal relationships under different reference schemes or montage designs~\cite{qin2010comparative,chella2016impact}. This mismatch also affects task transfer. Motor imagery emphasizes sensorimotor rhythm modulation~\cite{pfurtscheller1999event}, mental arithmetic involves workload-related oscillatory patterns~\cite{chikhi2022eeg}, and clinical detection tasks focus on abnormal waveforms, event-specific patterns, or ictal activity~\cite{gemein2020machine,obeid2016temple, fisher2014epileptic}. A representation tied to the source channels may entangle task-relevant neural patterns with dataset-specific spatial regularities, limiting transfer to different EEG paradigms~\cite{lotte2018review,wu2022transfer}.

A fixed region-token space provides a more consistent representation under montage variation. Individual electrodes measure local scalp potentials and remain sensitive to displacement and local noise~\cite{nielsen2023electrode, miljevic2022eegconnectivity}. Task-related EEG patterns often involve several electrodes~\cite{schaworonkow2022sensor}, and region tokens consolidate this distributed evidence without changing the representation structure when the channel configuration changes. This regional organization still requires selective aggregation. Uniform pooling cannot represent non-uniform channel contributions to the same region (Figure~\ref{fig:idea}, Top), and many-to-one aggregation may attenuate local waveforms that support downstream prediction. HyBRA addresses these limitations with feature-dependent channel-to-region matching and a direct channel-evidence pathway (Figure~\ref{fig:idea}, Bottom).

We propose HyBRA, a self-supervised EEG framework based on hyperbolic global-local brain-region aggregation. HyBRA combines short-term channel features with electrode coordinates and maps the resulting channel tokens into the fixed region-token space with the Hyperbolic Region Mixer. A brain-region prior defines admissible channel-region associations, and feature-dependent hyperbolic distance adjusts the relative weight of each candidate channel. The resulting region-token space forms the global component of HyBRA.

The complementary channel-evidence pathway forms the local component by drawing on the pre-aggregation channel features and injecting weighted channel-level evidence into each region token. The fused region tokens enter a Region-Time Transformer that models temporal dynamics within regions and interactions among regions. The self-supervised objective reconstructs latent features and region-level signals at masked region-time positions.

HyBRA introduces a region-centered pretraining scheme in which the regional pathway constructs a fixed region-token space and the channel-evidence pathway preserves channel-level evidence. Hyperbolic matching adapts channel contributions to each region, and evidence injection limits information loss during aggregation. HyBRA achieves the highest mean balanced accuracy on each of six downstream datasets. Controlled comparisons on TUAB and TUEV show higher mean balanced accuracy for hyperbolic matching than for parameter-matched Euclidean and cosine variants.

\section{Related Work}
\label{sec:related_work}

Recent self-supervised EEG models, including BIOT~\cite{yang2023biot}, EEGPT~\cite{wang2024eegpt}, LaBraM~\cite{jiang2024labram}, CBraMod~\cite{wang2025cbramod}, and CSBrain~\cite{zhou2025csbrain}, learn transferable representations from large EEG corpora and support downstream adaptation. These models handle montage variation using architecture-specific input adaptation, flexible tokenization, or learned channel embeddings. A recent benchmark reports that channel adaptation remains sensitive to the model architecture and training regime~\cite{kokate2026channel}. HyBRA differs at the representation level by mapping the observed channels into a fixed region-token space before region-time modeling.

Region-aware EEG models have established regional organization as a useful spatial representation. LGGNet~\cite{ding2024lggnet} aggregates electrodes within local regions before modeling inter-region relations. R2G-STLT~\cite{cheng2024r2g} constructs regional patches for spatiotemporal learning. LAG~\cite{gao2024lag} derives higher-level topology from local functional connectivity. ALGGNet~\cite{wang2025alggnet} studies dataset-specific regional partitions using adaptive clustering. These studies support the value of regional modeling. HyBRA differs by refining prior-defined channel--region associations from the current channel features of each sample. Regional aggregation may also attenuate channel-level evidence, which HyBRA retains with a channel-evidence pathway.

Hyperbolic representation learning offers a non-Euclidean geometry for structured and non-uniform relations~\cite{nickel2017poincare, ganea2018hyperbolic}. HEEGNet~\cite{li2026heegnet} reports geometric properties of EEG signals and learned representations that are compatible with hyperbolic modeling. Other hyperbolic EEG studies operate on complete sample representations or preconstructed connectivity networks~\cite{chang2025multiscale, guillemaud2025hyperbolic}. HyBRA uses feature-dependent hyperbolic distance for channel-to-region matching within a brain-region prior. Its channel-evidence pathway preserves channel-level evidence during aggregation.

\section{Method}
\label{sec:method}
\begin{figure*}[t]
    \centering
    \includegraphics[width=\textwidth]{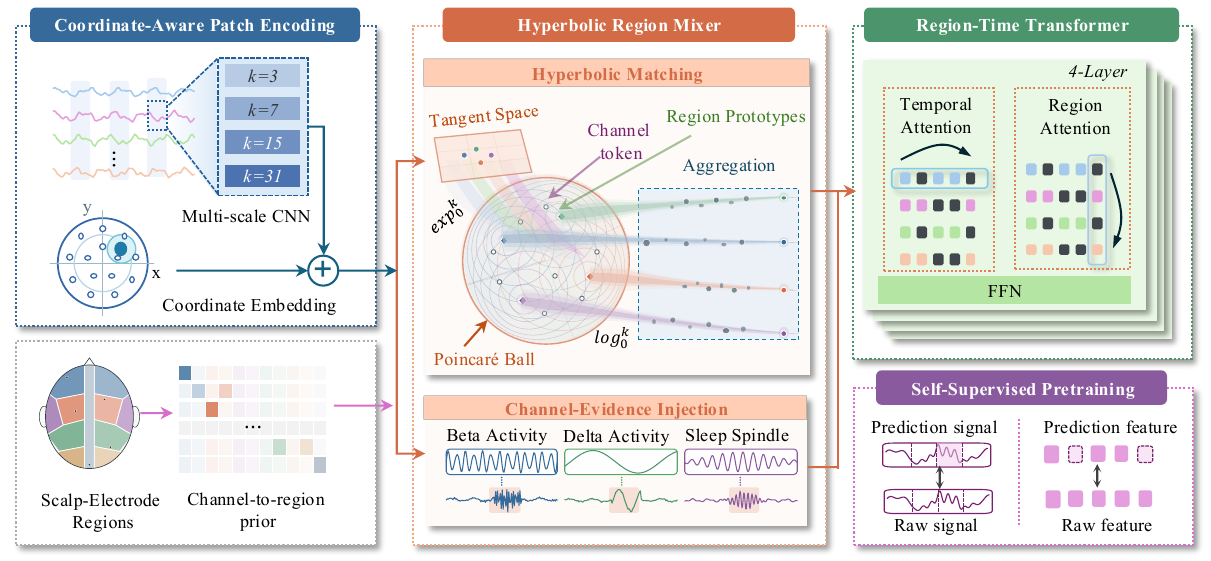}
    \caption{Overview of HyBRA-EEG. Coordinate-aware channel patches enter parallel prior-constrained hyperbolic matching and channel-evidence injection paths. Their sum forms fixed region tokens, which are modeled by the Region-Time Transformer. Self-supervised pretraining predicts masked latent features and reconstructs region-level signals.}
    \label{fig:framework}
\end{figure*}

HyBRA maps montage-dependent channel observations to a fixed region-token space while retaining informative channel evidence. Figure~\ref{fig:framework} presents this design through the functional blocks detailed below.

\subsection{Coordinate-Aware Patch Encoding}
\label{subsec:patch_encoding}

Given a batch of EEG segments $\mathbf{X}\in\mathbb{R}^{B\times C\times T}$, each channel is divided into $P$ temporal patches and processed by a shared temporal encoder:
\begin{equation}
    \mathbf{H}_{b,c,p}
    =
    f_{\theta}
    \left(
        \mathbf{X}^{\mathrm{p}}_{b,c,p}
    \right),
    \quad
    \mathbf{H}
    \in
    \mathbb{R}^{B\times C\times P\times D},
    \label{eq:patch_encoding}
\end{equation}
where $D$ is the feature dimension. We encode the three-dimensional electrode coordinate $\mathbf{q}_{c}\in\mathbb{R}^{3}$ obtained from the standard 10--05 electrode system using $\phi:\mathbb{R}^{3}\rightarrow\mathbb{R}^{D}$ and add it to every patch from channel $c$:
\begin{equation}
    \widetilde{\mathbf{H}}_{b,c,p}
    =
    \mathbf{H}_{b,c,p}
    +
    \phi
    \left(
        \mathbf{q}_{c}
    \right).
    \label{eq:coordinate_fusion}
\end{equation}
The coordinate-aware channel tokens $\widetilde{\mathbf{H}}$ are passed to the Hyperbolic Region Mixer. More details are in Appendix~A.

\subsection{Hyperbolic Region Mixer}
\label{subsec:hyperbolic_aggregation}

Given coordinate-aware channel tokens $\widetilde{\mathbf{H}}$, the Hyperbolic Region Mixer maps montage-dependent observations to $\mathbf{R}\in\mathbb{R}^{B\times K\times P\times D}$ over a fixed set of $K$ scalp-electrode regions. This shared region axis supports inter-region modeling across montages.

Let $\mathbf{A}\in[0,1]^{C\times K}$ denote the montage-specific
channel-to-region prior. For a nonempty region $k$,
$\mathcal{C}_{k}=\{c\mid A_{c,k}>0\}$ contains its candidate channels.
We map the channel tokens and learnable region prototypes to the
Poincar\'e ball:
\begin{equation}
\begin{aligned}
    \mathbf{z}_{b,c,p}
    &=
    \exp^{\kappa}_{\mathbf{0}}
    \left(
        \mathbf{W}_{h}\widetilde{\mathbf{h}}_{b,c,p}
        +
        \mathbf{b}_{h}
    \right),\\
    \boldsymbol{\rho}_{k}
    &=
    \exp^{\kappa}_{\mathbf{0}}
    \left(
        \boldsymbol{\mu}_{k}
    \right),\\
    d_{b,c,k,p}
    &=
    d_{\mathbb{B}}^{\kappa}
    \left(
        \mathbf{z}_{b,c,p},
        \boldsymbol{\rho}_{k}
    \right).
\end{aligned}
\label{eq:hyp_mapping_distance}
\end{equation}
Here, $\kappa>0$ is the magnitude of the fixed curvature $-\kappa$. The vectors $\mathbf{z}_{b,c,p}$ and $\boldsymbol{\rho}_{k}$ denote the hyperbolic channel embedding and region prototype. Let
$\ell_{c,k}=\log(\max(A_{c,k},\epsilon))$, with $\epsilon=10^{-8}$. For $\mathcal{C}_{k}\neq\varnothing$, the matching weights are
\begin{equation}
    \alpha_{b,c,k,p}
    =
    \operatorname{softmax}_{c\in\mathcal{C}_{k}}
    \bigl[
        \ell_{c,k}
        -
        \lambda d_{b,c,k,p}
    \bigr],
    \label{eq:hyp_aggregation_weight}
\end{equation}
where $\lambda=\operatorname{softplus}(\beta)>0$ scales the feature-dependent hyperbolic distance. Channels outside $\mathcal{C}_{k}$ receive zero weight. The prior sets the initial relative weights when its nonzero entries differ, and the distance adapts them to the current channel features. The embeddings are returned to the tangent space for aggregation:
\begin{equation}
    \mathbf{r}^{\mathrm{agg}}_{b,k,p}
    =
    \sum_{c=1}^{C}
    \alpha_{b,c,k,p}
    \log^{\kappa}_{\mathbf{0}}
    \left(
        \mathbf{z}_{b,c,p}
    \right).
    \label{eq:hyp_region_token}
\end{equation}
For an empty region, we set $\alpha_{b,c,k,p}=1/C$. The resulting tensor $\mathbf{r}^{\mathrm{agg}}\in\mathbb{R}^{B\times K\times P\times D}$ preserves the fixed region axis.

Many-to-one aggregation can attenuate channel-specific evidence. A channel-evidence injection branch draws directly from the pre-aggregation tokens. A normalized region-specific linear scorer gives
\begin{equation}
    s_{b,c,k,p}
    =
    \mathbf{w}_{k}^{\top}
    \operatorname{LN}
    \left(
        \widetilde{\mathbf{h}}_{b,c,p}
    \right).
    \label{eq:channel_content_score}
\end{equation}
For $\mathcal{C}_{k}\neq\varnothing$, the evidence weights and region evidence are
\begin{equation}
\begin{aligned}
    \eta_{b,c,k,p}
    &=
    \operatorname{softmax}_{c\in\mathcal{C}_{k}}
    \bigl[
        s_{b,c,k,p}
        +
        \ell_{c,k}
    \bigr],\\
    \mathbf{e}^{\mathrm{ch}}_{b,k,p}
    &=
    \sum_{c=1}^{C}
    \eta_{b,c,k,p}
    \widetilde{\mathbf{h}}_{b,c,p}.
\end{aligned}
\label{eq:channel_evidence}
\end{equation}
Channels outside $\mathcal{C}_{k}$ receive zero evidence weight, while an empty region uses $\eta_{b,c,k,p}=1/C$. The evidence is injected through a region-specific, feature-wise gate:
\begin{equation}
\begin{aligned}
    \mathbf{R}_{b,k,p}
    &=
    \mathbf{r}^{\mathrm{agg}}_{b,k,p}\\
    &\quad+
    \sigma\left(\boldsymbol{\gamma}_{k}\right)
    \odot
    \operatorname{Dropout}
    \Bigl(
        \mathbf{W}_{o}
        \operatorname{LN}
        \left(
            \mathbf{e}^{\mathrm{ch}}_{b,k,p}
        \right)
        +
        \mathbf{b}_{o}
    \Bigr).
\end{aligned}
\label{eq:final_region_representation}
\end{equation}
Here, $\boldsymbol{\gamma}_{k}\in\mathbb{R}^{D}$ controls the injected
evidence along each feature dimension, and
$\mathbf{b}_{o}\in\mathbb{R}^{D}$ is the bias of the shared output
projection. The final token combines hyperbolic region aggregation with
direct channel evidence. Appendix~B details the channel-to-region prior
$\mathbf{A}$, the empty-region fallback, and the Poincar\'e operations.

\subsection{Region-Time Transformer}
\label{subsec:region_time_transformer}

The Hyperbolic Region Mixer outputs region-time tokens $\mathbf{R}\in\mathbb{R}^{B\times K\times P\times D}$. The Region-Time Transformer operates on this grid. Temporal attention first updates the $P$ patches within each region. Region attention then exchanges information among the $K$ regions at each patch.

Learnable region and temporal embeddings are added before the first block:
\begin{equation}
    \mathbf{Z}^{0}_{b,k,p}
    =
    \mathbf{R}_{b,k,p}
    +
    \mathbf{e}^{r}_{k}
    +
    \mathbf{e}^{t}_{p},
    \label{eq:region_time_embedding}
\end{equation}
where $\mathbf{e}^{r}_{k},\mathbf{e}^{t}_{p}\in\mathbb{R}^{D}$.

In the $\ell$-th block, temporal attention is followed by region attention. Using pre-normalization and residual connections,
\begin{equation}
\begin{aligned}
    \mathbf{Z}^{\prime}
    &=
    \mathbf{Z}^{\ell}
    +
    \operatorname{Attn}_{\mathrm{time}}
    \left(
        \operatorname{LN}
        \left(
            \mathbf{Z}^{\ell}
        \right)
    \right), \\
    \mathbf{Z}^{\prime\prime}
    &=
    \mathbf{Z}^{\prime}
    +
    \operatorname{Attn}_{\mathrm{region}}
    \left(
        \operatorname{LN}
        \left(
            \mathbf{Z}^{\prime}
        \right)
    \right), \\
    \mathbf{Z}^{\ell+1}
    &=
    \mathbf{Z}^{\prime\prime}
    +
    \operatorname{FFN}
    \left(
        \operatorname{LN}
        \left(
            \mathbf{Z}^{\prime\prime}
        \right)
    \right).
\end{aligned}
\label{eq:region_time_block}
\end{equation}
Repeating the block $L$ times yields $\mathbf{Z}^{L}$ for self-supervised pretraining and downstream prediction. Architectural settings are given in Appendix~C.

\subsection{Self-Supervised Pretraining}
\label{subsec:self_supervised_pretraining}

HyBRA performs masked region-time modeling by sampling a binary mask
$\mathbf{M}\in\{0,1\}^{B\times K\times P}$ over the position-aware
tokens $\mathbf{Z}^{0}$ from Eq.~\ref{eq:region_time_embedding}. Each
masked token is replaced by a learnable token
$\mathbf{m}\in\mathbb{R}^{D}$:
\begin{equation}
    \widetilde{\mathbf{Z}}^{0}
    =
    (1-\mathbf{M})
    \odot
    \mathbf{Z}^{0}
    +
    \mathbf{M}
    \odot
    \mathbf{m}.
    \label{eq:masked_region_time_tokens}
\end{equation}
The Region-Time Transformer processes the corrupted region-time grid:
\[
\widetilde{\mathbf{Z}}^{L}
=
f_{\mathrm{RT}}(\widetilde{\mathbf{Z}}^{0}).
\]
Predictions are evaluated only at masked positions.

At each masked position, $\mathcal{L}_{\mathrm{feat}}$ is the cosine loss between the predicted feature and the pre-mask token at that position. The target is detached after LayerNorm and then $\ell_2$-normalized. $\mathcal{L}_{\mathrm{raw}}$ is the reconstruction loss for a normalized region-level signal derived from the input patches using the channel-to-region prior. The pretraining objective is
\begin{equation}
    \mathcal{L}_{\mathrm{SSL}}
    =
    \mathcal{L}_{\mathrm{feat}}
    +
    \lambda_{\mathrm{raw}}
    \mathcal{L}_{\mathrm{raw}},
    \label{eq:self_supervised_objective}
\end{equation}
where $\lambda_{\mathrm{raw}}$ weights signal reconstruction. Appendix~D gives the full pretraining procedure.

\section{Experiments}

\subsection{Experimental Setup}

\noindent\textbf{Datasets.}
Table~\ref{tab:datasets} summarizes the seven public subsets used for pretraining. They cover steady-state visual evoked potential (SSVEP)~\cite{liu2020beta,wang2017benchmark}, motor imagery (MI)~\cite{cho2017eeg,schalk2009eegmmidb}, and emotion recognition (EMO)~\cite{chen2023faced,li2019classification}. The montage and segment length vary across datasets. All signals are resampled to 200~Hz. 
The six downstream datasets cover motor imagery, clinical EEG analysis,
and mental workload. OpenBMI-MI~\cite{lee2019openbmi} uses six
subject-group folds, and BCIC-IV 2a~\cite{tangermann2012review} follows
a nine-fold leave-one-subject-out protocol. TUAB and
TUEV~\cite{obeid2016temple} use their official test sets.
Siena~\cite{detti2020eeg} uses 14 patient-level folds, while
MentalArithmetic~\cite{zyma2019mental} uses six subject-group folds.
All held-out test sets are subject- or patient-disjoint from the corresponding training data. Appendix~E describes the
preprocessing and split construction.

\begin{table}[t]
\centering
\small
\setlength{\tabcolsep}{2.6pt}
\renewcommand{\arraystretch}{1.05}
\begin{tabular}{cllrrc}
\toprule
\textbf{Phase} & \textbf{Dataset} & \textbf{Type} &
\textbf{\# Ch.} & \textbf{\# Samp.} & \textbf{Dur.} \\
\midrule
\multirow{7}{*}{\rotatebox{90}{\textbf{Pre-train}}}
& BETA (2s)       & SSVEP & 60 & 2,400   & 2s  \\
& BETA (3s)       & SSVEP & 60 & 8,800   & 3s  \\
& Tsinghua SSVEP  & SSVEP & 60 & 8,400   & 5s  \\
& Cho2017         & MI    & 64 & 10,520  & 3s  \\
& PhysioNet-MI    & MI    & 64 & 9,270   & 4s  \\
& FACED           & EMO   & 30 & 33,880  & 3s  \\
& SEED-V          & EMO   & 60 & 38,191  & 3s  \\
\midrule
\multirow{6}{*}{\rotatebox{90}{\textbf{Fine-tune}}}
& OpenBMI-MI      & MI    & 56 & 21,600  & 4s  \\
& BCIC-IV 2a      & MI    & 22 & 5,184   & 4s  \\
& TUAB            & ABN   & 19 & 409,757 & 10s \\
& TUEV            & EVT   & 19 & 113,353 & 5s  \\
& Siena           & SEIZ  & 19 & 50,499  & 10s \\
& MentalArith.    & MWL   & 19 & 1,707   & 5s  \\
\bottomrule
\end{tabular}
\caption{Datasets used for pretraining and downstream fine-tuning.
All signals are resampled to 200~Hz.}
\label{tab:datasets}
\end{table}

\begin{table*}[!t]
\centering
\small
\setlength{\tabcolsep}{2.7pt}
\renewcommand{\arraystretch}{1.05}
\begin{tabular}{lcccccc}
\toprule
\multirow{2}{*}{\textbf{Method}} &
\multicolumn{3}{c}{\textbf{TUAB}} &
\multicolumn{3}{c}{\textbf{TUEV}} \\
\cmidrule(lr){2-4}
\cmidrule(lr){5-7}
& \textbf{Balanced Acc.}
& \textbf{AUPRC}
& \textbf{AUROC}
& \textbf{Balanced Acc.}
& \textbf{Cohen's Kappa}
& \textbf{Weighted-F1} \\
\midrule
EEGNet
& 0.7642 $\pm$ 0.0036
& 0.8299 $\pm$ 0.0043
& 0.8412 $\pm$ 0.0031
& 0.3876 $\pm$ 0.0143
& 0.3577 $\pm$ 0.0155
& 0.6539 $\pm$ 0.0120 \\

EEGConformer
& 0.7758 $\pm$ 0.0049
& 0.8427 $\pm$ 0.0054
& 0.8445 $\pm$ 0.0038
& 0.4074 $\pm$ 0.0164
& \underline{0.3967 $\pm$ 0.0195}
& \underline{0.6983 $\pm$ 0.0152} \\

SPaRCNet
& \underline{0.7896 $\pm$ 0.0018}
& 0.8414 $\pm$ 0.0018
& \underline{0.8676 $\pm$ 0.0012}
& \underline{0.4161 $\pm$ 0.0262}
& \textbf{0.4233 $\pm$ 0.0181}
& \textbf{0.7024 $\pm$ 0.0104} \\

ContraWR
& 0.7746 $\pm$ 0.0041
& 0.8421 $\pm$ 0.0104
& 0.8456 $\pm$ 0.0074
& \textbf{0.4384 $\pm$ 0.0349}
& 0.3912 $\pm$ 0.0237
& 0.6893 $\pm$ 0.0136 \\

CNN-Transformer
& 0.7777 $\pm$ 0.0022
& 0.8433 $\pm$ 0.0039
& 0.8461 $\pm$ 0.0013
& 0.4087 $\pm$ 0.0161
& 0.3815 $\pm$ 0.0134
& 0.6854 $\pm$ 0.0293 \\

FFCL
& 0.7848 $\pm$ 0.0038
& \underline{0.8448 $\pm$ 0.0065}
& 0.8569 $\pm$ 0.0051
& 0.3979 $\pm$ 0.0104
& 0.3732 $\pm$ 0.0188
& 0.6783 $\pm$ 0.0120 \\

ST-Transformer
& \textbf{0.7966 $\pm$ 0.0023}
& \textbf{0.8521 $\pm$ 0.0026}
& \textbf{0.8707 $\pm$ 0.0019}
& 0.3984 $\pm$ 0.0228
& 0.3765 $\pm$ 0.0306
& 0.6823 $\pm$ 0.0190 \\
\midrule
BIOT
& 0.7959 $\pm$ 0.0057
& 0.8792 $\pm$ 0.0023
& 0.8815 $\pm$ 0.0043
& 0.5281 $\pm$ 0.0225
& 0.5273 $\pm$ 0.0249
& 0.7492 $\pm$ 0.0082 \\

EEGPT
& 0.7983 $\pm$ 0.0030
& --
& 0.8718 $\pm$ 0.0050
& 0.6232 $\pm$ 0.0114
& 0.6351 $\pm$ 0.0134
& 0.8187 $\pm$ 0.0063 \\

LaBraM
& 0.8140 $\pm$ 0.0019
& 0.8965 $\pm$ 0.0016
& \textbf{0.9022 $\pm$ 0.0009}
& 0.6409 $\pm$ 0.0065
& 0.6637 $\pm$ 0.0093
& 0.8312 $\pm$ 0.0052 \\

CSBrain
& \underline{0.8172 $\pm$ 0.0043}
& \underline{0.9005 $\pm$ 0.0066}
& 0.8957 $\pm$ 0.0046
& \underline{0.6903 $\pm$ 0.0059}
& \underline{0.6833 $\pm$ 0.0047}
& \underline{0.8333 $\pm$ 0.0057} \\

HyBRA
& \textbf{0.8224 $\pm$ 0.0021}
& \textbf{0.9082 $\pm$ 0.0029}
& \underline{0.9012 $\pm$ 0.0020}
& \textbf{0.7140 $\pm$ 0.0312}
& \textbf{0.6894 $\pm$ 0.0158}
& \textbf{0.8383 $\pm$ 0.0070} \\
\midrule[0.8pt]

\multirow{2}{*}{\textbf{Method}} &
\multicolumn{3}{c}{\textbf{BCIC-IV 2a}} &
\multicolumn{3}{c}{\textbf{OpenBMI-MI}} \\
\cmidrule(lr){2-4}
\cmidrule(lr){5-7}
& \textbf{Balanced Acc.}
& \textbf{Weighted-F1}
& \textbf{Cohen's Kappa}
& \textbf{Balanced Acc.}
& \textbf{AUPRC}
& \textbf{AUROC} \\
\midrule
HyBRA
& \textbf{0.5233 $\pm$ 0.1117}
& \textbf{0.5071 $\pm$ 0.1240}
& \textbf{0.3645 $\pm$ 0.1489}
& \textbf{0.8117 $\pm$ 0.0436}
& \textbf{0.8952 $\pm$ 0.0362}
& \textbf{0.8959 $\pm$ 0.0362} \\

CSBrain
& \underline{0.5054 $\pm$ 0.0710}
& \underline{0.4891 $\pm$ 0.0849}
& \underline{0.3405 $\pm$ 0.0947}
& \underline{0.8104 $\pm$ 0.0351}
& \underline{0.8429 $\pm$ 0.0492}
& \underline{0.8775 $\pm$ 0.0351} \\

LaBraM
& 0.4468 $\pm$ 0.1021
& 0.4234 $\pm$ 0.1142
& 0.2623 $\pm$ 0.1361
& 0.7013 $\pm$ 0.0284
& 0.7580 $\pm$ 0.0518
& 0.7723 $\pm$ 0.0326 \\

EEGPT
& 0.4419 $\pm$ 0.0703
& 0.4177 $\pm$ 0.0949
& 0.2559 $\pm$ 0.0938
& 0.6810 $\pm$ 0.0354
& 0.7505 $\pm$ 0.0535
& 0.7558 $\pm$ 0.0491 \\

BIOT
& 0.4277 $\pm$ 0.1338
& 0.4025 $\pm$ 0.1465
& 0.2369 $\pm$ 0.1784
& 0.6908 $\pm$ 0.0318
& 0.7656 $\pm$ 0.0585
& 0.7721 $\pm$ 0.0421 \\
\midrule[0.8pt]

\multirow{2}{*}{\textbf{Method}} &
\multicolumn{3}{c}{\textbf{Siena}} &
\multicolumn{3}{c}{\textbf{MentalArithmetic}} \\
\cmidrule(lr){2-4}
\cmidrule(lr){5-7}
& \textbf{Balanced Acc.}
& \textbf{Weighted-F1}
& \textbf{AUROC}
& \textbf{Balanced Acc.}
& \textbf{AUPRC}
& \textbf{AUROC} \\
\midrule
HyBRA
& \textbf{0.7291 $\pm$ 0.1158}
& \textbf{0.8404 $\pm$ 0.1489}
& 0.7950 $\pm$ 0.1147
& \textbf{0.7663 $\pm$ 0.0501}
& \textbf{0.7047 $\pm$ 0.1025}
& \textbf{0.8662 $\pm$ 0.0376} \\

LaBraM
& \underline{0.7262 $\pm$ 0.1314}
& 0.8031 $\pm$ 0.2195
& \textbf{0.8175 $\pm$ 0.1775}
& \underline{0.6964 $\pm$ 0.0508}
& \underline{0.5563 $\pm$ 0.0946}
& \underline{0.7683 $\pm$ 0.0552} \\

BIOT
& 0.7036 $\pm$ 0.1060
& \underline{0.8152 $\pm$ 0.1815}
& \underline{0.8054 $\pm$ 0.1064}
& 0.6416 $\pm$ 0.0703
& 0.4602 $\pm$ 0.1154
& 0.7019 $\pm$ 0.1052 \\

CSBrain
& 0.6807 $\pm$ 0.1135
& 0.7790 $\pm$ 0.1592
& 0.7421 $\pm$ 0.1228
& 0.6538 $\pm$ 0.0964
& 0.4691 $\pm$ 0.1374
& 0.7213 $\pm$ 0.1103 \\

EEGPT
& 0.5960 $\pm$ 0.0863
& 0.7575 $\pm$ 0.1431
& 0.6627 $\pm$ 0.1602
& 0.5512 $\pm$ 0.0368
& 0.3147 $\pm$ 0.0526
& 0.5666 $\pm$ 0.0556 \\
\bottomrule
\end{tabular}
\caption{Downstream results on six EEG datasets, reported as mean $\pm$ standard deviation. For TUAB and TUEV, task-specific baselines~\cite{lawhern2018eegnet,song2023eegconformer,jing2023development,yang2023contrawr,peh2022transformer,li2022ffcl,song2021sttransformer} and EEG foundation models form separate comparison groups; for the remaining datasets, all EEG foundation models are compared together. Bold and underlined values denote the best and second-best results within each group.}
\label{tab:downstream_results}
\end{table*}

\noindent\textbf{Pretraining Setup.}
HyBRA is pretrained for 200 epochs on eight Huawei Ascend 910B3 NPUs with a batch size of 32 per device. AdamW uses a peak learning rate of $3\times10^{-4}$ and a weight decay of $1\times10^{-4}$. The learning rate increases linearly for five epochs and follows a cosine schedule to $1\times10^{-6}$. The mask ratio is 0.4, and $\lambda_{\mathrm{raw}}$ is 0.5. Appendix~F lists the software versions and remaining settings.

\noindent\textbf{Downstream Setup.}
Pretrained EEG models are fine-tuned end to end on each downstream dataset. Within a dataset, all methods use the same splits, preprocessing, training budget, and validation criterion. Each backbone uses its required input adapter and downstream head. HyBRA passes the complete region-time token grid $\mathbf{Z}^{L}$ to a compact constrained CNN head. It is optimized with AdamW using a learning rate of $1\times10^{-4}$ and a weight decay of $1\times10^{-4}$. The checkpoint with the best validation result is evaluated once on the held-out test set. Dataset-specific settings are in Appendix~G.

\subsection{Main Results}
\label{subsec:main_results}

\noindent\textbf{Baselines and Evaluation Metrics.}
We compare HyBRA with four pretrained EEG models: BIOT, EEGPT, LaBraM, and CSBrain. TUAB and TUEV also include task-specific baselines. Balanced accuracy (BAcc) is the primary metric as it averages recall across classes. Secondary metrics follow each dataset's evaluation setting and are listed in Table~\ref{tab:downstream_results}.

\begin{table}[!t]
\centering
\fontsize{9}{10.5}\selectfont
\begin{tabular}{@{}l@{\hspace{2.5pt}}c@{\hspace{2.5pt}}c@{\hspace{5pt}}c@{\hspace{2.5pt}}c@{}}
\toprule
& \multicolumn{2}{c@{\hspace{5pt}}}{OpenBMI-MI}
& \multicolumn{2}{c@{}}{MentalArithmetic} \\
\cmidrule(lr){2-3}
\cmidrule(l){4-5}
Method & BAcc & Params & BAcc & Params \\
\midrule
\textbf{HyBRA}
& \textbf{81.17$\pm$4.36}
& \textbf{1.45 (1.0$\times$)}
& \textbf{76.63$\pm$5.01}
& \textbf{1.45 (1.0$\times$)} \\

CSBrain
& 81.04$\pm$3.51
& 44.86 (30.9$\times$)
& 65.38$\pm$9.64
& 28.06 (19.3$\times$) \\

LaBraM
& 70.13$\pm$2.84
& 5.82 (4.0$\times$)
& 69.64$\pm$5.08
& 5.82 (4.0$\times$) \\

BIOT
& 69.08$\pm$3.18
& 3.19 (2.2$\times$)
& 64.16$\pm$7.03
& 3.19 (2.2$\times$) \\

EEGPT
& 68.10$\pm$3.54
& 25.32 (17.4$\times$)
& 55.12$\pm$3.68
& 25.32 (17.4$\times$) \\
\bottomrule
\end{tabular}
\caption{BAcc (\%) and parameter counts on OpenBMI-MI and MentalArithmetic. BAcc is reported as mean $\pm$ standard deviation
over six subject-group folds. Parameters are in millions, with ratios relative to HyBRA in parentheses. Bold values indicate the highest BAcc or smallest model.}
\label{tab:model_efficiency}
\end{table}

\begin{table}[!t]
\centering
\fontsize{9}{10.5}\selectfont
\setlength{\tabcolsep}{3pt}
\renewcommand{\arraystretch}{1.08}
\begin{tabular}{@{}llccc@{}}
\toprule
& & \multicolumn{3}{c}{Number of brain regions} \\
\cmidrule(l){3-5}
Dataset & Metric & 5 & 10 & 11 \\
\midrule

\multirow{2}{*}{TUAB}
& BAcc (\%)
& 81.82$\pm$0.57
& 81.17$\pm$0.74
& \textbf{82.24$\pm$0.21} \\

& Kappa (\%)
& 63.59$\pm$1.17
& 62.34$\pm$1.51
& \textbf{64.38$\pm$0.45} \\

\midrule

\multirow{2}{*}{OpenBMI-MI}
& BAcc (\%)
& 72.25$\pm$4.82
& 80.90$\pm$4.32
& \textbf{81.17$\pm$4.36} \\

& Kappa (\%)
& 44.51$\pm$9.64
& 61.81$\pm$8.64
& \textbf{62.33$\pm$8.71} \\

\bottomrule
\end{tabular}
\caption{Effect of brain-region granularity on TUAB and OpenBMI-MI. The 5-, 10-, and 11-region partitions use coarse lobar groups, bilateral lobar groups, and bilateral groups with a midline token, respectively. Results are mean $\pm$ standard deviation over five random seeds for TUAB and six subject-group folds for OpenBMI-MI. Bold values indicate the best result.}
\label{tab:region_number}
\end{table}

\noindent\textbf{Performance Comparison.}
Table~\ref{tab:downstream_results} shows that HyBRA achieves the highest mean BAcc on all six downstream datasets. Its macro-averaged BAcc is 0.7278, giving relative improvements of 5.03\%, 8.48\%, 15.29\%, and 18.29\% over the corresponding macro averages of CSBrain, LaBraM, BIOT, and EEGPT. Against the strongest pretrained baseline on each dataset, the BAcc gains are 2.37, 1.79, and 6.99 percentage points on TUEV, BCIC-IV 2a, and MentalArithmetic. The margins on TUAB, OpenBMI-MI, and Siena are smaller but remain positive. On OpenBMI-MI, HyBRA obtains a comparable BAcc to CSBrain ($81.17\%$ versus $81.04\%$) and improves AUROC and AUPRC by 1.84 and 5.23 percentage points. HyBRA also leads every reported metric on TUEV, BCIC-IV 2a, and MentalArithmetic. LaBraM retains the highest AUROC on TUAB and Siena. The BAcc results show gains across both clinical and BCI datasets without implying uniform improvements on every secondary metric.

\begin{table*}[!t]
\centering
\small
\setlength{\tabcolsep}{8pt}
\renewcommand{\arraystretch}{1.08}
\begin{tabular}{llcc}
\toprule
\textbf{Module} & \textbf{Setup} & \textbf{TUAB} & \textbf{TUEV} \\
\midrule

Coordinate-Aware Patch Encoding
& w/o coordinate encoding
& 0.8057 $\pm$ 0.0051
& 0.6555 $\pm$ 0.0340 \\

\midrule

\multirow{4}{*}{Hyperbolic Region Mixer}
& Euclidean distance-based matching
& 0.8134 $\pm$ 0.0078
& 0.6509 $\pm$ 0.0214 \\

& Cosine similarity-based matching
& 0.8164 $\pm$ 0.0035
& 0.5989 $\pm$ 0.0206 \\

& w/o brain-region tokenization
& 0.7980 $\pm$ 0.0073
& 0.6166 $\pm$ 0.0175 \\

& w/o channel evidence injection
& 0.8123 $\pm$ 0.0049
& 0.6468 $\pm$ 0.0428 \\

\midrule

\multirow{2}{*}{Region-Time Transformer}
& region-only attention
& 0.8137 $\pm$ 0.0028
& 0.6814 $\pm$ 0.0089 \\

& temporal-only attention
& 0.8081 $\pm$ 0.0078
& 0.6469 $\pm$ 0.0270 \\

\midrule

\multirow{3}{*}{Self-Supervised Pretraining}
& w/o SSL pretraining
& 0.8123 $\pm$ 0.0060
& 0.6704 $\pm$ 0.0468 \\

& w/o latent feature prediction
& 0.8130 $\pm$ 0.0038
& 0.6758 $\pm$ 0.0347 \\

& w/o region-level signal reconstruction
& 0.8122 $\pm$ 0.0054
& 0.6158 $\pm$ 0.0129 \\

\midrule

Full Model
& HyBRA
& \textbf{0.8224 $\pm$ 0.0021}
& \textbf{0.7140 $\pm$ 0.0312} \\

\bottomrule
\end{tabular}
\caption{Ablation BAcc on TUAB and TUEV, reported as mean $\pm$ standard deviation over five seeds. The Euclidean and cosine variants replace only the matching metric. The variant without brain-region tokenization applies the same axial Transformer and SSL objectives directly to channel-time tokens, while the variant without SSL pretraining trains HyBRA from random initialization. Bold values indicate the best result.}
\label{tab:ablation}
\end{table*}

\subsection{Model Size and Brain-Region Granularity}
\label{subsec:modeff_analysis}

\noindent\textbf{Model Size.}
Table~\ref{tab:model_efficiency} compares downstream parameter counts and BAcc on OpenBMI-MI and MentalArithmetic. HyBRA uses 1.45M parameters in both settings and achieves the highest mean BAcc. On OpenBMI-MI, its BAcc is $81.17\%$, compared with $81.04\%$ for CSBrain, whose downstream model contains 30.9 times as many parameters. On MentalArithmetic, HyBRA raises the next-best BAcc from $69.64\%$ to $76.63\%$ while using one quarter of the parameters of LaBraM. These results show that the performance gain does not arise from increased model size.

\noindent\textbf{Brain-Region Granularity.}
Table~\ref{tab:region_number} compares three brain-region partitions. The 5-region partition retains coarse lobar organization. The 10-region partition separates the left and right hemispheres. The 11-region partition further assigns electrodes marked with \textit{z} to a dedicated midline token (\textit{MID}), following standardized EEG nomenclature~\cite{seeck2017standardized}.

On TUAB, the 10-region partition performs below the 5-region partition. Adding the midline token gives the highest BAcc of $82.24\pm0.21\%$ and Cohen's kappa of $64.38\pm0.45\%$. On OpenBMI-MI, hemispheric separation raises BAcc from $72.25\pm4.82\%$ to $80.90\pm4.32\%$. The 11-region partition further increases BAcc to $81.17\pm4.36\%$. Cohen's kappa follows the same pattern. The comparison supports the selected 11-region partition and shows that token organization matters beyond the number of regions.

\begin{figure}[!t]
    \centering
    \includegraphics[width=\columnwidth]
    {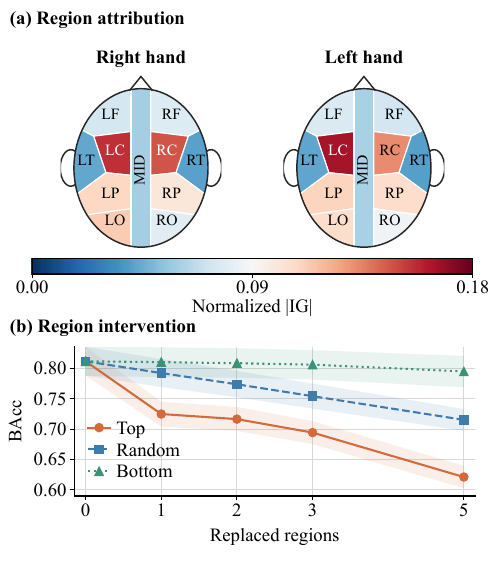}
    \caption{Region attribution and token perturbation on OpenBMI-MI. (a) Subject-averaged normalized absolute Integrated Gradients (IG) attribution for each motor imagery class. (b) Subject-level BAcc after replacing the most-attributed, randomly selected, or least-attributed region tokens in each trial with the corresponding fold-specific training means. Shaded bands indicate bootstrap 95\% confidence intervals across 54 subjects.}
    \label{fig:interpretability}
\end{figure}

\begin{figure*}[!t]
    \centering
    \includegraphics[width=\textwidth]{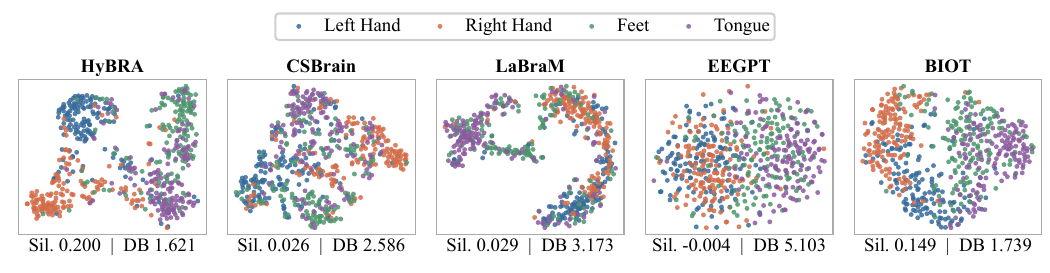}
    \caption{t-SNE visualization of downstream representations on BCIC-IV 2a. Each point denotes one EEG trial, and colors indicate motor-imagery classes. Each panel reports the Silhouette coefficient and Davies--Bouldin index computed on the displayed two-dimensional embedding.}
    \label{fig:tsne}
\end{figure*}

\subsection{Ablation Study}
\label{subsec:ablation_study}

We evaluate the components of HyBRA on TUAB and TUEV using the same data splits, random seeds, and training protocols. Table~\ref{tab:ablation} shows that every variant has a lower mean BAcc than the full model. The larger reductions on TUEV suggest greater sensitivity to spatial modeling and the pretraining objective under this event-classification setting.

\noindent\textbf{Coordinate-Aware Patch Encoding.}
Removing coordinate encoding lowers BAcc by 1.67 percentage points on TUAB and 5.85 points on TUEV. Electrode coordinates provide useful spatial information before channel-to-region matching.

\noindent\textbf{Hyperbolic Region Mixer.}
Replacing hyperbolic matching with parameter-matched Euclidean or cosine variants lowers mean BAcc on both datasets. The variant without brain-region tokenization is a structural tokenization ablation and is not parameter-matched to HyBRA. It removes region aggregation, channel-evidence injection, and region embeddings, and applies the same axial Transformer and SSL objectives to the channel-time grid with per-channel reconstruction. This reduces BAcc by 2.44 and 9.74 percentage points on TUAB and TUEV, respectively. Removing channel-evidence injection reduces BAcc by 1.01 and 6.72 points, supporting fixed region tokenization and the complementary evidence pathway.

\noindent\textbf{Region-Time Transformer.}
Using only region attention or temporal attention lowers performance on both datasets. The complete Region-Time Transformer performs best, showing that temporal modeling within each region and interaction across regions provide complementary information.

\noindent\textbf{Self-Supervised Pretraining.}
Training the complete architecture from random initialization lowers BAcc by 1.01 points on TUAB and 4.36 points on TUEV. Removing latent feature prediction also reduces performance on both datasets. Removing region-level signal reconstruction causes a larger decrease on TUEV, from 0.7140 to 0.6158. The complete pretraining objective achieves the highest mean BAcc on both datasets.

\subsection{Interpretability Analysis}
\label{subsec:interpretability}

\noindent\textbf{Region Attribution and Perturbation.}
We examine whether the region tokens produced by the Hyperbolic Region Mixer carry differentiated evidence for downstream predictions. Under the channel-to-region prior, the left-central (LC) and right-central (RC) tokens cover central scalp areas~\cite{pfurtscheller2001motor}. For right- and left-hand imagery, LC and RC jointly account for 30.27\% and 29.74\% of the normalized absolute Integrated Gradients attribution~\cite{sundararajan2017axiomatic}, respectively, as shown in Figure~\ref{fig:interpretability}(a). This non-uniform pattern shows that the region tokens contribute differently to the prediction. The concentration of attribution in central scalp regions is spatially consistent with established sensorimotor patterns in hand motor imagery.

As shown in Figure~\ref{fig:interpretability}(b), we further test attribution faithfulness through token perturbation~\cite{samek2017evaluating}. The model is frozen, and the selected region tokens in each trial are replaced with their fold-specific training-set means. Replacing the highest-attributed token reduces mean subject-level BAcc from 0.8117 to 0.7248. Random and lowest-attributed replacement retain BAcc values of 0.7923 and 0.8102, respectively. The decrease caused by highest-attributed replacement is 6.75 percentage points larger than that caused by random replacement. Replacing the five highest-attributed tokens further reduces BAcc to 0.6211, while replacing the five lowest-attributed tokens retains 0.7949. These results show that model predictions are more sensitive to highly attributed region tokens, supporting the faithfulness of the attribution ranking. Results under alternative attribution targets are reported in Appendix~H.


\noindent\textbf{Representation Space Visualization.}
We use t-SNE~\cite{maaten2008tsne} to compare the representations learned by the pretrained EEG models on BCIC-IV 2a. Figure~\ref{fig:tsne} shows that the HyBRA embedding forms more compact within-class groups and clearer separation between motor-imagery classes. The other embeddings exhibit greater class mixing, although some clustering remains visible for BIOT. HyBRA also obtains the highest Silhouette coefficient of 0.200 and the lowest Davies--Bouldin index of 1.621 among the displayed representations. These visualization-based results are consistent with the stronger downstream performance of HyBRA on BCIC-IV 2a, while not serving as independent evidence of class separability in the original high-dimensional space.

\section{Conclusion}
\label{sec:conclusion}

We presented HyBRA, a self-supervised EEG framework that maps montage-dependent channel observations to a fixed brain-region token space. Its hyperbolic global-local brain-region aggregation combines a montage-specific region prior with feature-dependent hyperbolic distance, while channel-evidence injection preserves information from the original channel tokens. HyBRA achieves the highest mean BAcc across six downstream datasets. The Euclidean and cosine variants on TUAB and TUEV further support the contribution of hyperbolic channel-to-region matching under the evaluated settings. Future work will examine adaptive region construction and broader pretraining across centers and electrode configurations.

\bibliography{aaai2027}


\end{document}